\documentclass[conference]{IEEEtran}
\IEEEoverridecommandlockouts
\usepackage{cite}
\usepackage{amsmath,amssymb,amsfonts}
\usepackage{algorithmic}
\usepackage{graphicx}
\usepackage{textcomp}
\usepackage{xcolor}
\def\BibTeX{{\rm B\kern-.05em{\sc i\kern-.025em b}\kern-.08em
    T\kern-.1667em\lower.7ex\hbox{E}\kern-.125emX}}

\begin{document}
    
    \title{Facilitating Human-Robot Collaboration through Natural Vocal Conversations\\
    }
    
    \author{\IEEEauthorblockN{Davide Ferrari, Filippo Alberi and Cristian Secchi}
    \IEEEauthorblockA{\textit{Department of Sciences and Methods of Engineering} \\
    \textit{University of Modena and Reggio Emilia}\\
    Reggio Emilia, Italy\\
    \{davide.ferrari95, cristian.secchi\}@unimore.it}
    }
    
    \maketitle
    
    \begin{abstract}
    
        In the rapidly evolving landscape of human-robot collaboration, effective communication between humans and robots is crucial for complex task execution. Traditional request-response systems often lack naturalness and may hinder efficiency. In this study, we propose a novel approach that employs human-like conversational interactions for vocal communication between human operators and robots. The framework emphasizes the establishment of a natural and interactive dialogue, enabling human operators to engage in vocal conversations with robots. Through a comparative experiment, we demonstrate the efficacy of our approach in enhancing task performance and collaboration efficiency. The robot's ability to engage in meaningful vocal conversations enables it to seek clarification, provide status updates, and ask for assistance when required, leading to improved coordination and a smoother workflow.
        The results indicate that the adoption of human-like conversational interactions positively influences the human-robot collaborative dynamic. Human operators find it easier to convey complex instructions and preferences, fostering a more productive and satisfying collaboration experience.
        
    \end{abstract}

    \begin{IEEEkeywords}
        Human-Robot Communication, Vocal Conversation
    \end{IEEEkeywords}

    \section{Introduction}

    The growing integration of robots in the industrial domain has given rise to a more complex landscape of collaboration between humans and machines \cite{Baratta2023PCS}. This new era of human-robot interaction presents numerous opportunities to enhance efficiency and productivity in work activities. However, it also calls for a profound reflection on optimizing the communicative processes among the involved actors \cite{Mukherjee2022Frontiers}. Notably, communication between humans and robots plays a fundamental role in determining the success of these collaborative interactions.
    
    The objective of this article is to explore and analyze a crucial aspect of human-robot collaboration: natural vocal communication \cite{Vargas2021Frontiers}. Traditionally, communication with robots has been based on request-response systems, where human operators issued specific commands and robots responded by executing the requested action \cite{MAVRIDIS201522}. In some works \cite{ferrari2023hfr}, voice is integrated with other modes of communication, while in others, it is structured to enable the robot to initiate the dialogue \cite{ferrari2022icra}. However, this mode of interaction can be limiting and inefficient when facing complex and dynamic tasks. Our work focuses on an innovative approach that introduces a higher level of interaction, akin to human-to-human communication, enabling natural vocal conversations between human operators and robots. Through the integration of elements of human language comprehension and context awareness into the robot's communication system, we aim to make the interaction more intuitive, flexible, and responsive.
    
    Throughout this article, we will present our framework for natural vocal communication and elucidate its implementation details. Additionally, we will conduct a comparative experiment to assess the efficacy of our approach, demonstrating how natural vocal conversations can enhance the robot's ability to understand human instructions and respond appropriately. The main contributions of this work are:
    \begin{itemize}
        \item Development of an innovative framework for natural vocal communication between humans and robots, based on interactive conversations similar to human interactions.
        \item Implementation of a comparative experiment to assess the effectiveness of the proposed framework.
    \end{itemize}

    \section{Natural Vocal Conversation Architecture}

    Consider a task of Human-Robot Collaboration (HRC) where both the robot and the human operator need to perform a complex task that requires shared knowledge, such as collaborative assembly. The operator and the robot must be able to communicate efficiently and exchange information about the task's progress, the need for tools or components, the requirement for assistance or support and so on. By utilizing deep learning approaches, it has become possible to establish a natural communication between humans and robots, allowing the creation of natural dialogues, similar to those that occur during human-to-human conversations, instead of the standard request-response interactions.

\begin{figure}[htbp]
        \centering
        \includegraphics[width=0.8\linewidth]{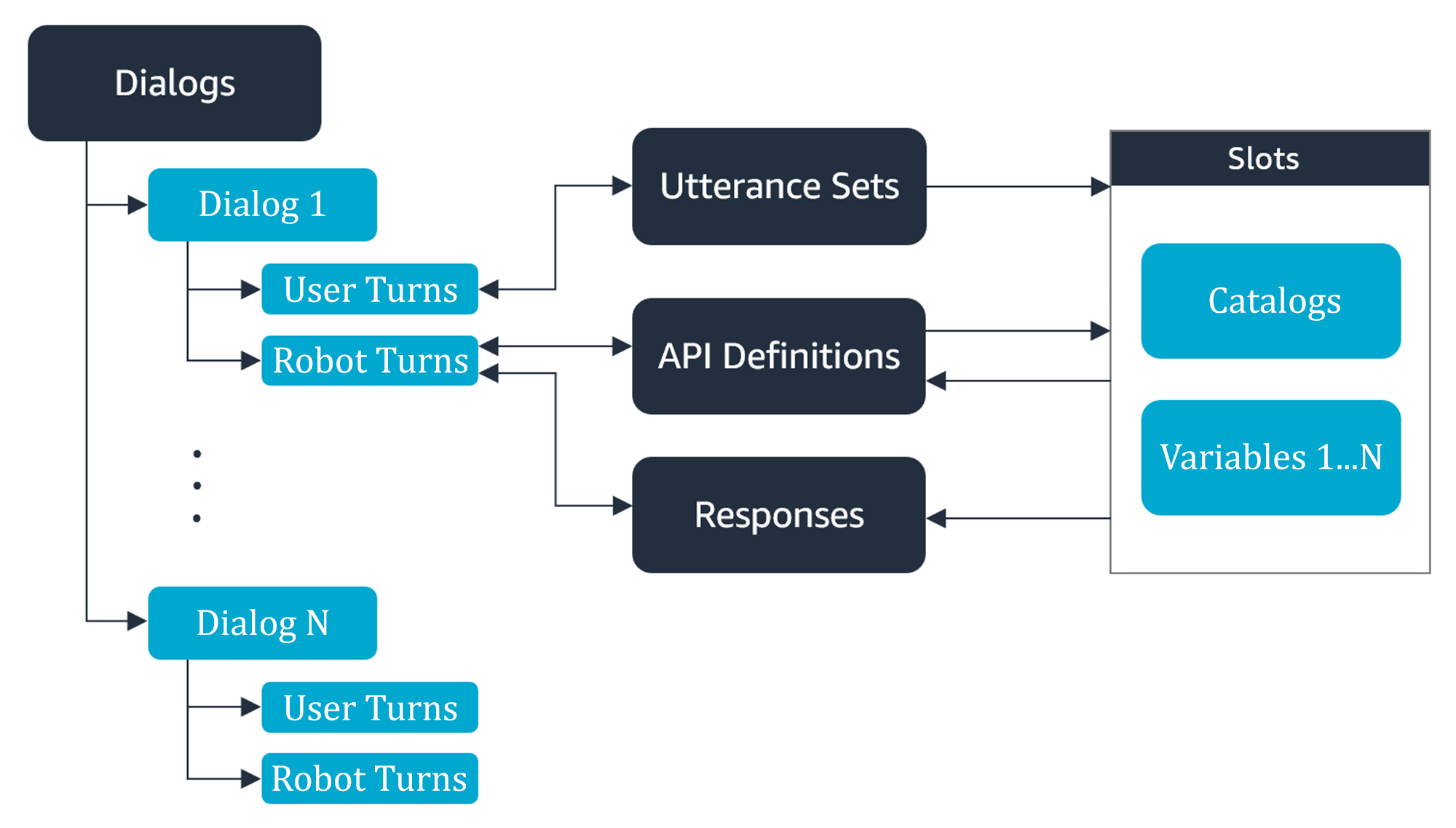}
        \caption{Conversation Structure}
        \label{fig:Conversation-Structure}
    \end{figure}

    \noindent
    The proposed architecture involves integrating a commercial voice assistant into a Human-Robot Collaboration job, enabling the robot and the operator to exchange information reciprocally through the created voice communication channel, structured in the form of conversations. Each conversation, as depicted in Figure \ref{fig:Conversation-Structure}, consists of a series of turn-taking dialogues, starting from the user's turn and then proceeding to the robot's turn. In each dialogue, the user's requests will be linked to one of the example phrases (utterance sets) defined during the construction of the dialogue schema, incorporating personalized variables or catalogs if present; the robot can simply respond, ask to full-fill any missing variable slots, or send the message to the back-end using a JSON-Request. Subsequently, based on the request, it can call an API that may refer to external objects or trigger other dialogues to continue the conversation with the human.

    \vspace{-1mm}

    \section{Experimental Validation}

    The experimental validation
    was conducted through a comparative experiment, simulating a collaborative assembly job where the operator had to assemble 5 components together in collaboration with a UR10e collaborative manipulator. The parts and tools were arranged in 3 areas with different access permissions (robot-only / human and robot), and they were needed to proceed with the final component assembly. Using the voice communication channel, the user could request tools, components or assistance, and the robot, through dialogues, could provide details, communicate issues, propose alternatives, or offer suggestions.
    The experiment aimed to compare the proposed conversation architecture with the standard request-response structure, by comparing the execution times and the number of correctly completed assembly steps. Additionally, a NASA-TLX \cite{NASA1988HART} questionnaire was used to gather user experiences during the task. The experiment was conducted on a sample of 20 participants, equally divided between the two experiments, to avoid any possible learning effect.

    \subsection{Implementation Details}
    
\begin{figure}[htbp]
            \centering
            \includegraphics[width=0.8\linewidth]{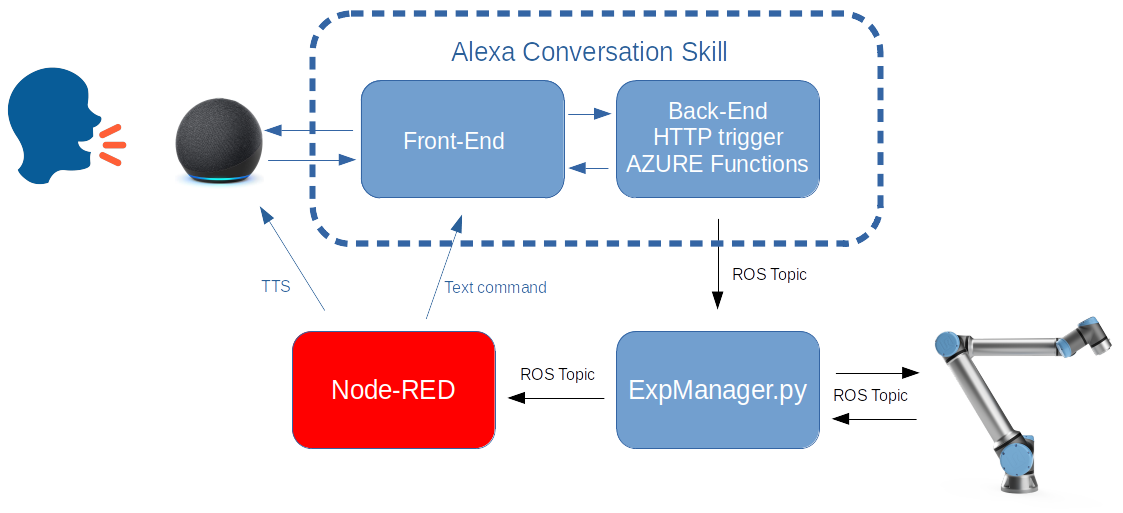}
            \caption{Architecture Scheme}
            \label{fig:Architecture-Scheme}
        \end{figure}

        The architecture in Figure \ref{fig:Architecture-Scheme} was developed by integrating a custom Amazon Alexa voice assistant skill into the ROS framework \cite{ros}. The skill was built using Alexa Conversations \cite{acharya2021alexa}, a Deep Learning-based approach that utilizes API calls to manage a multi-turn dialogue system between Alexa and the user, enabling more natural and human-like interactions. The skill's back-end was developed locally (non-Alexa-hosted) in Python, allowing seamless integration with ROS by leveraging Microsoft Azure's HTTP Trigger Functions. Additionally, to add Text-To-Speech functionality, Node-RED \cite{nodered} was incorporated, a web service for logical path programming that provides direct interaction with Alexa APIs, enabling the robot to report errors, handle events, and initiate a conversation by invoking specific dialogue APIs.

    \subsection{Analysis of the Results}

\begin{figure}[htbp]
            \minipage{0.24\textwidth}
                \includegraphics[trim={0 0 0 0}, clip, width=\linewidth]{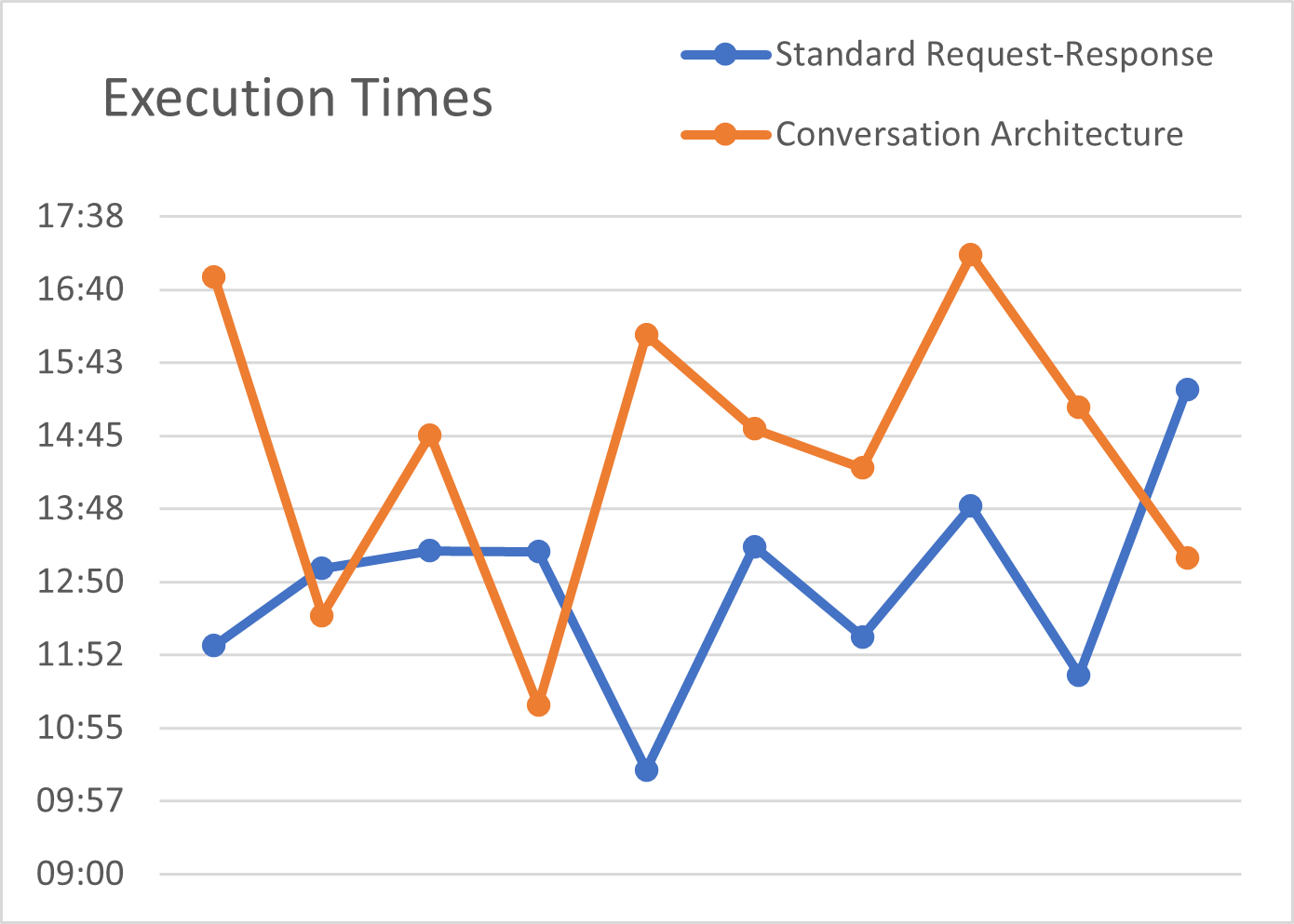}
            \endminipage\hfill
            \minipage{0.24\textwidth}
                \includegraphics[trim={0 0 0 0}, clip, width=\linewidth]{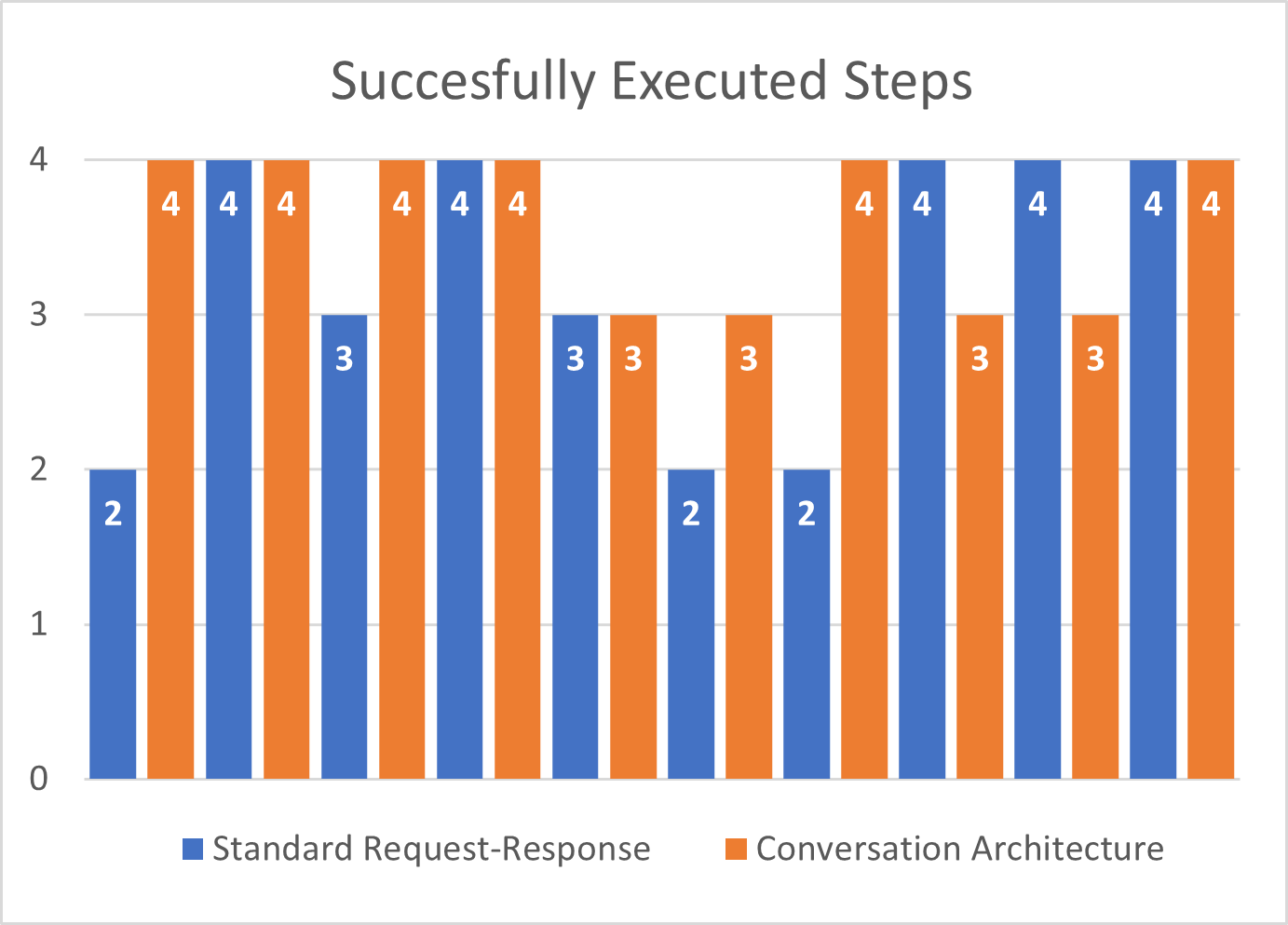}
            \endminipage\hfill
            \caption{Execution Times and Successfully-Completed Steps}
            \label{fig:results-time-steps}
        \end{figure}

        The obtained results have highlighted significant differences in the usage of the two architectures. The experiment execution times, graphed in Figure \ref{fig:results-time-steps}, were found to be approximately 12\% longer using the proposed architecture; this was an expected outcome since the incorporation of a more elaborate dialogue form lengthens the interactions compared to simple Request-Response commands. On the other hand, the use of Conversations showed a higher average precision in executing various steps, with an increase of approximately 10\%. Furthermore, the NASA-TLX analyses revealed a clear preference for user experience when using the proposed architecture, with a score of 57.67 compared to the standard's score of 55.

    \bibliographystyle{IEEEtran}
    \bibliography{Bibliography.bib}
    
\end{document}